\documentclass[9pt,twocolumn]{extarticle}
\usepackage[a4paper, textwidth=180mm, textheight=250mm, top=18mm,
  columnsep=6mm, footskip=8mm]{geometry}

\usepackage[T1]{fontenc}
\usepackage[utf8]{inputenc}
\usepackage[english]{babel}
\usepackage{amsmath,amssymb,amsthm}
\usepackage{newtxtext}
\usepackage{newtxmath}
\ifdefined\XeTeXversion
  \usepackage{fontspec}
\else
  \usepackage[scaled=0.92]{helvet}
\fi
\usepackage[scaled=0.95]{zlmtt}
\usepackage{microtype}
\usepackage[dvipsnames]{xcolor}
\usepackage{graphicx}
\usepackage{booktabs}
\usepackage{colortbl}
\usepackage{tabularx}
\usepackage{longtable}
\usepackage{array}
\usepackage{enumitem}
\usepackage{url}
\usepackage{xurl}  
\usepackage[super,sort&compress,comma]{natbib}
\usepackage{caption}
\usepackage{titlesec}
\usepackage{fancyhdr}
\usepackage[hidelinks]{hyperref}
\hypersetup{colorlinks=true, linkcolor=black, citecolor=black, urlcolor={blue!70!black}}

\definecolor{methodcolor}{HTML}{B03A2E}

\titleformat{\section}{\sffamily\bfseries\fontsize{11}{13}\selectfont}{}{0pt}{}
\titlespacing*{\section}{0pt}{10pt plus 2pt minus 2pt}{3pt}
\titleformat{\subsection}{\sffamily\bfseries\fontsize{9.5}{11.5}\selectfont}{}{0pt}{}
\titlespacing*{\subsection}{0pt}{8pt plus 2pt minus 2pt}{2pt}
\titleformat{\paragraph}[runin]{\bfseries}{}{0pt}{}
\titlespacing*{\paragraph}{0pt}{3pt plus 1pt}{0.5em}

\DeclareCaptionLabelSeparator{nature}{\,\textbar\ }
\addto\captionsenglish{}  

\newcommand{\firstpagenote}[1]{{\let\thefootnote\relax\footnotetext{#1}}}

\setlist{nosep, leftmargin=*}

\usepackage{marvosym}  
\IfFileExists{fontawesome6.sty}{\usepackage{fontawesome6}}{\usepackage{fontawesome5}}  
\usepackage{mathtools}
\usepackage{bm}
\usepackage{algorithm}
\usepackage{algorithmic}   
\usepackage{multirow}
\usepackage{float}
\usepackage{placeins}   
\usepackage{tikz}
\usetikzlibrary{arrows.meta,positioning,calc,fit}
\usepackage{cleveref}

\theoremstyle{plain}

\theoremstyle{definition}

\theoremstyle{remark}

\crefname{figure}{Fig.}{Figs.}\Crefname{figure}{Fig.}{Figs.}
\crefname{table}{Table}{Tables}\Crefname{table}{Table}{Tables}
\crefname{equation}{equation}{equations}\Crefname{equation}{Equation}{Equations}
\creflabelformat{equation}{(#2#1#3)}
\crefname{algorithm}{Algorithm}{Algorithms}\Crefname{algorithm}{Algorithm}{Algorithms}

\usepackage{adjustbox}      
\usepackage{verbatim}       
\usepackage[most]{tcolorbox} 
\usepackage{listings}       
\lstdefinestyle{jsonstyle}{
  basicstyle=\ttfamily\footnotesize, breaklines=true, breakatwhitespace=false, frame=single,
  columns=fullflexible, keepspaces=true, showstringspaces=false,
  literate={"}{{{\ttfamily\char34}}}1}

\newcommand{\datasetURL}{\url{https://huggingface.co/datasets/jiayijdong/VerifiedLab}}
\newcommand{\codeURL}{\url{https://github.com/jiayijdong/VerifiedLab}}

\makeatletter
\newcommand{\xr@lookup}[2]{%
  \@ifundefined{xr@#1@#2}%
    {\textbf{??}\@latex@warning{Cross-document reference `#2' undefined}}%
    {\@nameuse{xr@#1@#2}}}
\newcommand{\sref}[1]{\xr@lookup{S}{#1}}
\newcommand{\mref}[1]{\xr@lookup{M}{#1}}
\newcommand{\xr@def}[3]{\expandafter\gdef\csname xr@#1@#2\endcsname{#3}}
\makeatother

\input{supp_labels}
\definecolor{youtubered}{HTML}{FF0000}
\definecolor{hfyellow}{HTML}{FFD21E}
\definecolor{githubblack}{HTML}{24292E}
\newcommand{\aff}[1]{\raisebox{0.6ex}{\fontsize{6.3}{7}\selectfont #1}}
\newcommand{\papertitle}{VerifiedLab: Harnessing Knowns for Unknowns}
\newcommand{\runningtitle}{Yao et al.\ \textbar\ Harnessing Knowns for Unknowns}
\newcommand{\paperauthors}{%
  Yu Yao\aff{1,\dag,\Letter}, Jiayi Dong\aff{2,\dag}, Yang Yang\aff{2}, Shuiwang Ji\aff{3},
 Ju Li\aff{1,\Letter}}
\newcommand{\paperaffiliations}{%
  \aff{1}Massachusetts Institute of Technology, Cambridge, MA, USA. \aff{2}University of California, Los Angeles, CA, USA.
  \aff{3}Texas A\&M University, College Station, TX, USA.
  \aff{\dag}These authors contributed equally. \aff{\Letter}e-mail: yu\_yao@mit.edu; liju@mit.edu}
\newcommand{\papersidebar}{%
  \textbf{Correspondence}\\ yu\_yao@mit.edu\\ liju@mit.edu\\[6pt]
  \textbf{Code and Data}\\ 
  \href{https://github.com/jiayijdong/VerifiedLab}
  {\textcolor{githubblack}{\faGithub} VerifiedLab/Code}\\[6pt]
  \href{https://huggingface.co/datasets/jiayijdong/VerifiedLab}
  {\raisebox{-1.5pt}{\hspace{-1pt}\includegraphics[height=9pt]{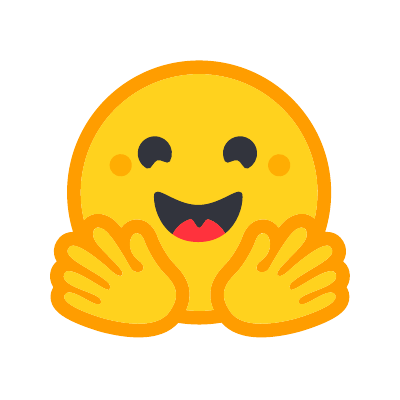}} VerifiedLab/Data}\\[6pt]
  \href{https://youtu.be/9TP1gdmzB4I?si=xZFMLxit33f8pfj6}
  {\textcolor{youtubered}{\faIcon{youtube}} VerifiedLab/Demo}\\[6pt]
}

\graphicspath{{figures/}}
\fancypagestyle{plain}{\fancyhf{}\fancyfoot[R]{\sffamily\footnotesize\bfseries\thepage}}

\begin{document}

\twocolumn[{%
\begin{flushleft}
\includegraphics[height=14pt]{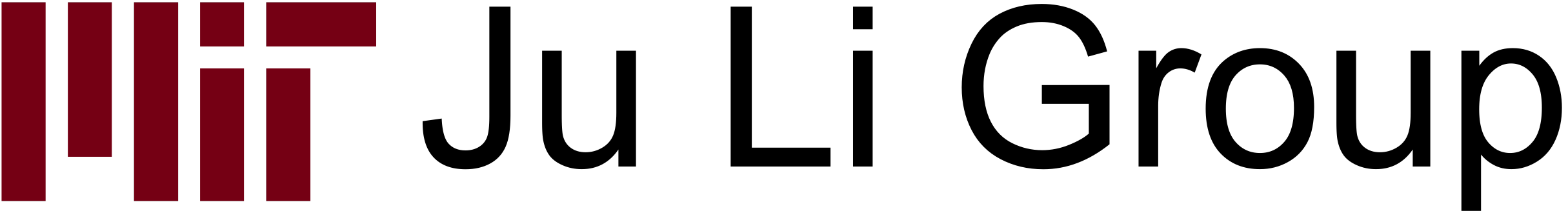}\par
\vspace{3pt}\hrule height 0.6pt\vspace{8pt}
{\sffamily\bfseries\fontsize{21}{24}\selectfont \papertitle\par}
\vspace{8pt}\hrule height 0.4pt\vspace{8pt}
\begin{minipage}[t]{0.24\textwidth}
\sffamily\footnotesize\raggedright \papersidebar
\end{minipage}\hfill
\begin{minipage}[t]{0.73\textwidth}
{\sffamily\bfseries\fontsize{9.5}{12}\selectfont\raggedright \paperauthors\par}
\vspace{9pt}
{\fontsize{10.5}{13.6}\selectfont
Across human history, knowledge has undergone several transitions, from orality to writing and then to digital archives. At each transition, humans established new conventions to preserve knowledge for the new era. Nowadays, as research agents are increasingly deployed for laboratory experiments, knowledge is crossing another boundary: from records written for humans to those for machines. In this work, we introduce VerifiedLab, a multi-agent harness framework for this human-to-machine transition of knowledge. VerifiedLab transforms research output from \emph{passive artifacts} into \emph{active systems} that accelerate downstream use, adoption and prediction of future trends. In VerifiedLab, a team of AI scientists with distinct roles reasons over records verified field by field: a principal investigator plans, a librarian computes the evidence directly from the records, specialists draft, and a gate and a critic check each claim against the measured reliability of its data before the harness decides whether to release or withhold it. We demonstrate the efficacy of VerifiedLab through extensive experiments on perovskite solar cells: VerifiedLab recovers known physical trends, anticipates devices reported a year later, and lets AI scientists inherit the literature together with a measure of how far it can be trusted. By scaling the inference-time verification compute, VerifiedLab introduces a new pathway for knowledge dissemination and machine harness for AI co-scientists.\par}
\end{minipage}
\end{flushleft}
\vspace{10pt}
}]

\firstpagenote{\footnotesize\paperaffiliations}


\noindent For most of human history, knowledge was carried in human minds. How to knap a flint, read the weather or set a broken bone was held in memory, shown by hand and learned by watching. Across generations, social learning allowed knowledge to accumulate beyond individual discovery\cite{Boyd2011CulturalNiche}. But a chain of minds is a fragile medium. Each time a skill passed from one mind to another, a step could be dropped. An instruction could be misremembered. A recipe could be repeated without the circumstance that had made it work. Preserving improvements through such handovers is central to cumulative culture\cite{Dean2014CumulativeCulture,Mesoudi2018CumulativeCulturalEvolution}.

Writing gave knowledge a durable medium. A record on clay, papyrus or paper outlived its author, travelled beyond its author's reach, and could be consulted unchanged a century later. Printing multiplied the copies and lowered the cost of spreading them. More scholars in different cities could consult and compare copies of the same text\cite{Eisenstein1979PrintingPress}. Journals such as the seventeenth-century \emph{Philosophical Transactions} organized a continuing public record. Over subsequent centuries, conventions of authorship, methods reporting, editorial review and citation developed around that record\cite{Fyfe2022HistoryScientificJournals}. Each of these steps carried knowledge across a boundary: from mind to page, from manuscript to press, from private letter to public record. At each boundary, meaning could leak away until a convention was invented to hold it in place. A manuscript without an author could not be questioned, an experiment without a method could not be repeated, and a claim without a citation could not be traced. Science became trustworthy because of these conventions, not only because of the medium.

\begin{figure}[!t]
    \centering
    \includegraphics[width=\linewidth]{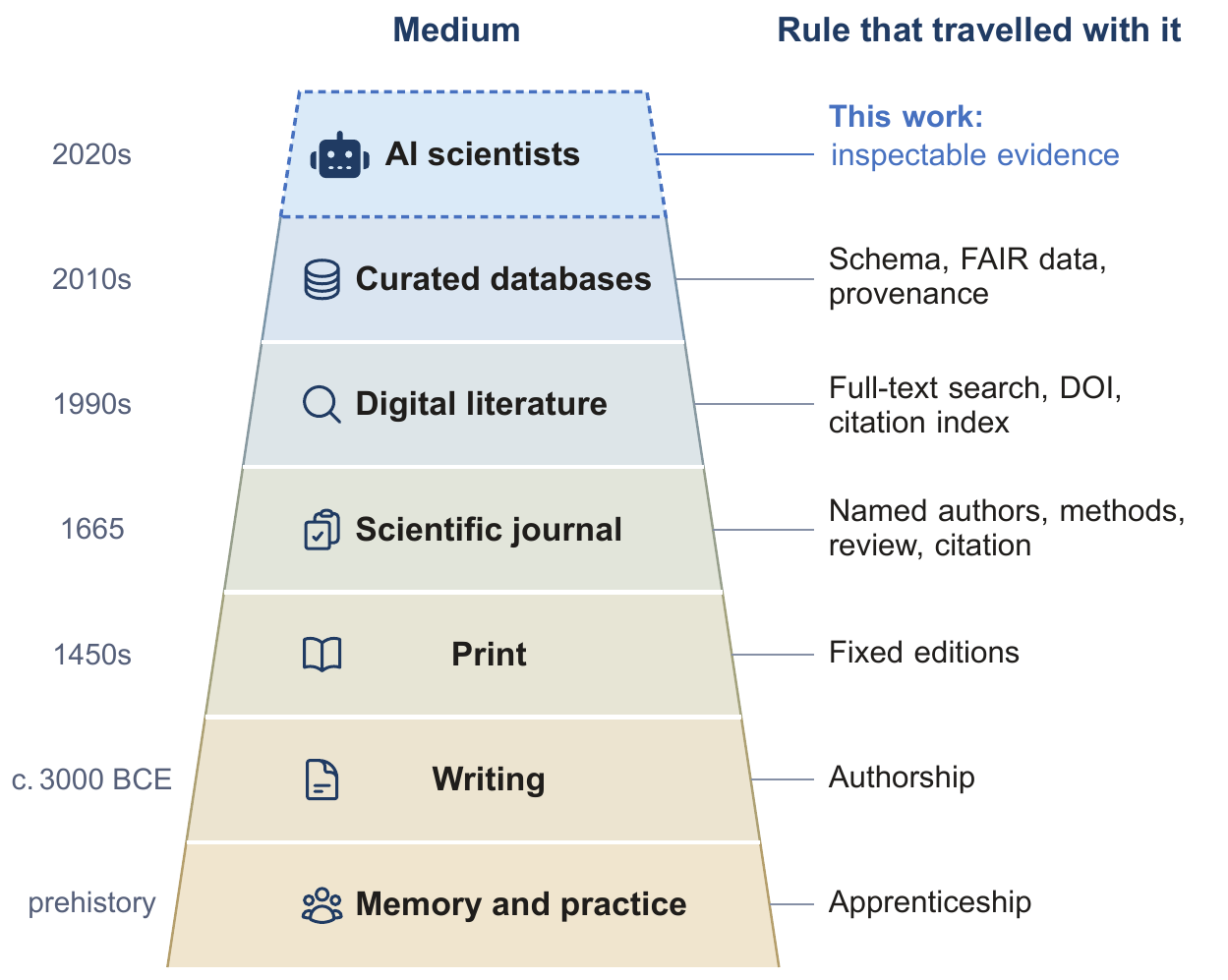}
    \captionsetup{font=footnotesize}
    \caption{\textbf{Scientific knowledge has crossed a boundary each time it changed medium.}
    Left, the media that carried knowledge; right, the rule that made each trustworthy. The apex is the boundary being crossed now, where AI scientists read papers and this work supplies the rule. Icons: Heroicons (MIT), Font Awesome Free (CC BY 4.0).}
    \label{fig:knowledge}
\end{figure}

For the literature, the present is a moment of déjà vu: it is crossing such a boundary again (Fig.~\ref{fig:knowledge}). Papers are now digital and searchable, while FAIR principles make the findability, provenance and reuse of scientific data explicit\cite{Wilkinson2016FAIR}. More and more of these records are read by machines, and the transition this time runs from papers written for human readers to structured records that machines act on. Research agents built on large language models (LLMs) read prior work, form hypotheses, plan and run experiments and choose the next experiment\cite{Lu2024AIScientist,Yamada2025AIScientistV2,Gottweis2025AICoScientist,Swanson2025VirtualLab,Miao2026Paper2Agent,Boiko2023Coscientist,Szymanski2023ALab}. Where experiments are themselves code, an agent can write a script, run it and read the answer\cite{Lu2024AIScientist,Huang2024MLAgentBench,Chan2025MLEBench}. In chemistry and materials science the situation is different. The knowledge an agent needs is spread over tens of thousands of papers. Each paper mixes prose, tables, figures and supporting files. Turning those reports into structured data is a separate information-extraction task\cite{Dagdelen2024StructuredExtraction,Xu2024LLM_IE_Survey}. Consider perovskite solar cells (PSCs), whose efficiency has advanced from early sensitized devices to certified research cells\cite{Kojima2009JACS,NREL_PV_Efficiency_Chart_2025}. Progress depends on composition\cite{Jeon2015NatureCompositional,Saliba2016ScienceRb}, interfaces\cite{Zhou2014ScienceInterface,Tan2022NatureHeterointerfaces}, processing chemistry\cite{Meng2024NatEnergyDMAFo,Wang2025ScienceSolventSystems}, tandem integration\cite{Li2025ScienceVaporPartitioning} and grain-boundary control\cite{He2026NatEnergyAmorphousGB}. The paper describes several devices; its best efficiency is in a table, the recipe for the light-absorbing layer in the methods, and the stability test in a supplementary figure with its own temperature, humidity and light. A solvent ratio mentioned in passing applies to one variant only. Each number belongs to one device, one step and one set of conditions; measurement and stability-reporting guidelines require those links for meaningful comparisons\cite{Christians2015Measurements,Khenkin2020ConsensusStabilityPSC}. A language model can recover the numbers while losing these links. It writes plausible wrong values: a reversed ratio, a layer attached to the wrong device, a stability figure separated from its protocol. Such a record passes a glance and fails in use, and its errors pass into every plan built on top of it. The alternative, curating the literature by hand, takes years of expert labour\cite{jacobsson2022perovskitedatabase}. Automated databases have captured device properties and selected fabrication details\cite{beard2022perovskite,valencia2025fabricationdb}; we extend this line of work to the complete record of each experiment, verified field by field.

This work addresses this challenge by converting each paper into a verified record that functions as a machine-readable account of the experiment, exposing its devices, procedures and measurements as fields that carry their evidence and their measured reliability. Language models extract materials data from scientific prose\cite{Dagdelen2024StructuredExtraction,Polak2024ChatExtract,Gupta2024PolymerExtraction}; text-mining systems recover chemical entities and synthesis descriptions\cite{Swain2016ChemDataExtractor,Kim2017MaterialsSynthesisTextMining}, while literature embeddings capture associations between materials\cite{Tshitoyan2019Mat2Vec}. Long-context benchmarks nevertheless show that access to a document does not ensure reliable use of evidence throughout it\cite{Liu2024LostMiddle}. Studies of LLM judges measure agreement with people and expose consistency and correctness limits\cite{Zheng2023LLMJudge,Tan2025JudgeBench,Lee2024ConsistencyLLMEvaluators}. Research-agent benchmarks assess experimentation, data-driven discovery and paper reproduction\cite{Huang2024MLAgentBench,Majumder2025DiscoveryBench,Chen2025ScienceAgentBench,Starace2025PaperBench,Siegel2026COREBench}. Scientific fact-checking and claim--evidence benchmarks assess support for claims, including evidence dispersed across full papers\cite{Wadden2020SciFact,Javaji2025ClaimBench}; attribution and citation benchmarks test the support of generated statements by identified sources\cite{Rashkin2023Attribution,Gao2023ALCE}. Each of these addresses a part of the task. We preserve device, procedure and measurement identity across an entire experimental record under field-specific scientific rules, with verifiers calibrated against experts and reliability estimates that stay with the records.

\begin{figure*}[!t]
    \centering
    \includegraphics[width=\linewidth]{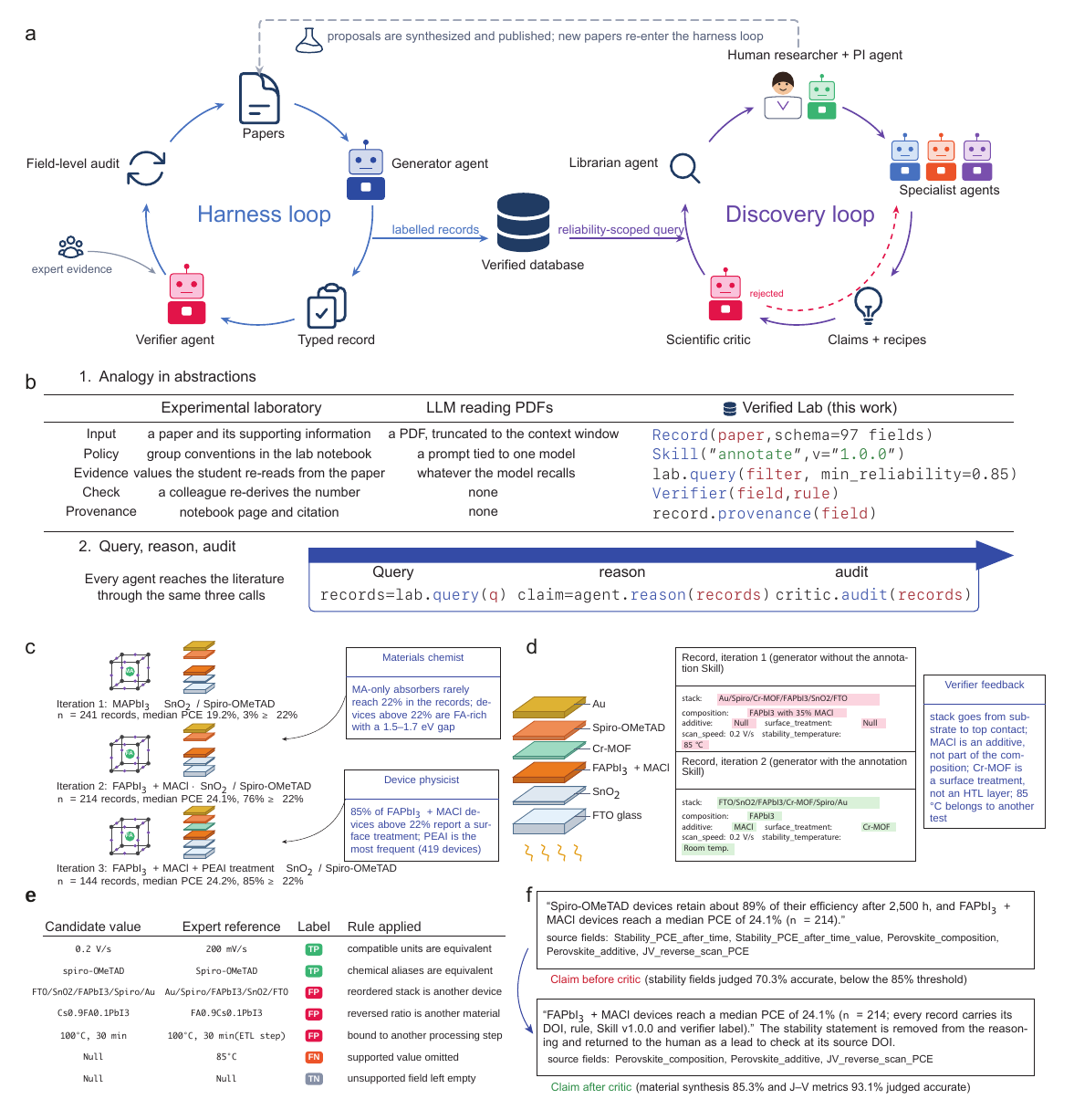}
    \captionsetup{font=footnotesize}
    \caption{\textbf{Overview of the verified-literature harness.}
    \textbf{a}, Two coupled agent loops. In the harness loop (blue), a generator agent fills one record per paper under the annotation Skill $A^{\alpha}$, a verifier agent labels every field under the verification Skill $V^{\beta}$, and the audit re-versions both Skills. In the discovery loop (purple), a PI agent, a librarian and specialist agents reason over the verified database, and a critic rejects claims without provenance (dashed red). Published proposals re-enter the harness as papers (grey).
    \textbf{b}, The objects that a human laboratory, an LLM reading PDFs and the Verified Lab use for input, policy, evidence, checking and provenance.
    \textbf{c}, A recipe refined over three iterations with specialist feedback; $n$ and the share above 22\% are computed over the matching verified records.
    \textbf{d}, A real record reconstructed without and with the annotation Skill (red, errors; green, released record), with the verifier's feedback.
    \textbf{e}, Verifier decisions on candidate--reference pairs and the rule applied.
    \textbf{f}, A claim before and after the critic. Icons: Heroicons (MIT), Font Awesome Free (CC BY 4.0).}
    \label{fig:overview}
\end{figure*}

In this work, we introduce VerifiedLab, a multi-agent harness framework for this human-to-machine transition of knowledge (Fig.~\ref{fig:overview}). We built the harness for perovskite solar cells and tested it at each stage, from the reading of a paper to the use of the resulting records in scientific reasoning. Frontier models recover most of a record but lose its procedures: current agents misplace values that are tied to several procedural steps in experimental materials science and chemistry, and they share the same failure patterns. Written rules close much of this gap on the verifier side, bringing verifiers to expert-level agreement regardless of model size. Run over the recent literature of the field, the harness returns a database in which every category of field carries a judged accuracy. On this database we place a team of AI scientists with distinct roles. A principal investigator plans the study, a librarian computes the evidence directly from the records, specialists draft the claims, and a gate and a critic check each claim against the measured reliability of its data before the harness decides whether to release or withhold it. Reasoning only over the evidence, the team recovers known physical trends and anticipates devices reported a year later, and it withholds the claims, on stability for instance, whose data are not yet reliable enough to carry them.

\subsection*{Verified Harness: Record, Policy and Check}

The verified harness has three parts: a record, a policy and a check. The record is a schema; a generator agent fills it, and a verifier agent checks every filled value against evidence, both under written policies (Fig.~\ref{fig:overview}a and Methods). The schema was written by experimentalists. Its fields, nearly a hundred of them, describe a paper's best-performing device in seven groups: reference data, cell definition, device stack, material synthesis, device fabrication, current--voltage (J--V) metrics and stability metrics. Each field has a fixed name and a convention for its value. Units stay attached to numbers. Layer stacks are written in order, as in \texttt{TiO2/ZrO2/carbon}. \texttt{Null} has one meaning: the paper does not say.

The scientific policy lives in versioned Skills rather than in prompts tied to one model. This makes an important dependency of the pipeline explicit, a concern also identified in work on technical debt in machine-learning systems\cite{Sculley2015TechnicalDebt}. Rigour is written down there, in the reviewing consensus of experts, the equivalence rules of chemistry and the boundary conditions of physics, and applied to every field. It is not left to a model that cannot be inspected or corrected. The annotation Skill gives a generator the rule for choosing the champion device, the priority among sources of evidence, the alignment of layers with values, the assignment of each measurement to its device and the conditions for writing \texttt{Null}. It forbids filling a field from general knowledge or from a different device. The verification Skill defines, for a verifier, the equivalence of two values. Chemical aliases, compatible units, numerical tolerances and ordered sequences are accepted. Reversed ratios, reordered layers and values borrowed from another device are rejected. For each field the verifier sees the schema entry, the candidate value, the expert reference and the rule, and it returns one of five labels: the values agree, both are empty, the candidate is missing, or the candidate is unsupported or wrong. Decisions are made field by field, so each can be audited on its own, and a wrong rule is caught by the audit and fixed in the next version. The score of a generator never assumes that the automated judge is right, because the judge is itself measured against expert decisions. The reference for those measurements comes from two expert-curated collections of PSC papers (Methods). One was annotated by us against the full schema; the other was drawn from a community-curated database\cite{jacobsson2022perovskitedatabase}. Experts annotated independently and resolved disagreements before any automated verifier was run, separating annotation, adjudication and evaluation in line with established quality-management practice\cite{Klie2024AnnotationQuality}. Many fields are legitimately \texttt{Null}, so a record left entirely empty already scores well above zero. This empty-record baseline is the floor against which generators are read. Figure~\ref{fig:verifiedlab} shows the whole workflow in use by a team of agents that reaches the literature only through the harness; we return to it after the benchmarks.

\subsection*{Procedural Information Hampers Recovery}

\begin{figure*}[!t]
    \centering
    \includegraphics[width=\linewidth]{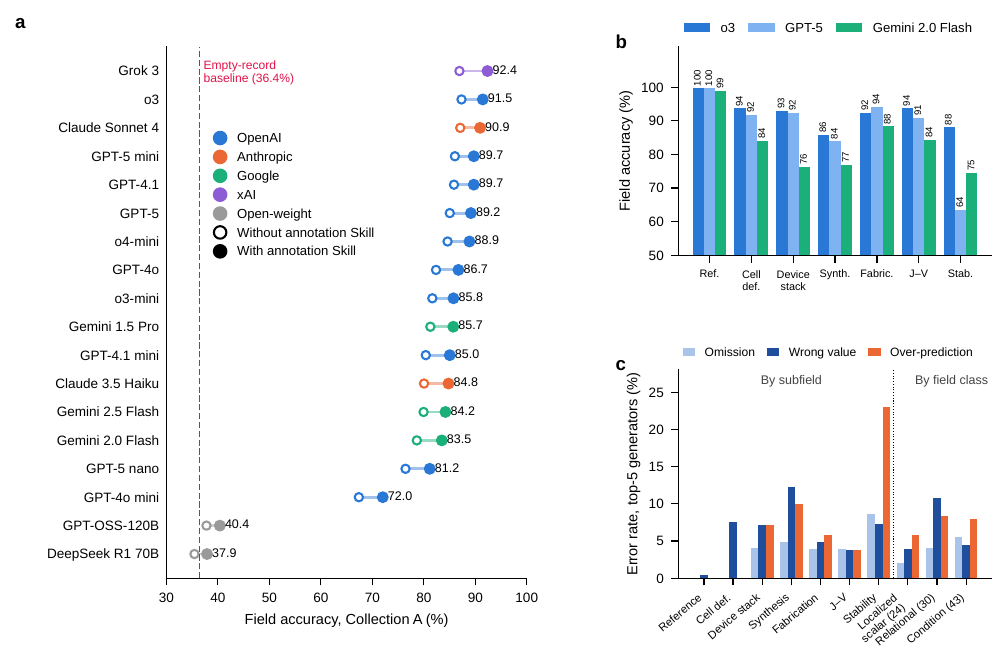}
    \caption{\textbf{Research agents reconstruct experimental records.}
    \textbf{a}, Field accuracy of 18 generators on Collection A (100 papers, 9,700 field decisions each) with (filled) and without (open) the annotation Skill; colours, model families; dashed line, empty-record baseline.
    \textbf{b}, Accuracy by subfield for three generators.
    \textbf{c}, Error composition for the five strongest generators, by subfield (left) and by the competence a field tests (right; localized scalars, 24 fields; relational structures, 30; conditions, 43). Omission, wrong value and over-prediction are defined in Methods. Decisions are those of the o3 verifier with the verification Skill.}
    \label{fig:generation}
\end{figure*}

The first test asks how much of a record today's readers recover, and which part of it they lose. We ran eighteen generators, proprietary and open-weight, through the fixed harness, so that differences between runs reflect the model alone. A calibrated verifier, described in the next section, scored them (Fig.~\ref{fig:generation}a and Supplementary Fig.~\sref{fig:ed-generators}). With the annotation Skill, the strongest proprietary generators recover most of each record. The two open-weight generators sit close to the empty-record floor, because most of their errors are fields they leave blank. The proprietary generators fail differently. For most of them, the largest class of error is a value that is present but wrong, ahead of omissions and ahead of values that the paper never gave.

The errors gather where the information is procedural (Fig.~\ref{fig:generation}b and Supplementary Fig.~\sref{fig:ed-subfields}). Bibliographic fields are at ceiling for every model. Stability and synthesis fields are where models part company, and in the stability fields the usual mistake is to invent a condition that the paper left unspecified (Fig.~\ref{fig:generation}c). To separate knowledge from binding, we grouped the fields by what each one asks of the reader. Localized scalars, which a paper states once in one place, are recovered best. Conditions tied to a single processing step or test come next. Relational fields, whose value is an ordered or multi-part structure attached to a layer or a device, are recovered worst, and they come back wrong rather than empty. A sizeable share of reported layer stacks and precursor-concentration lists do not match the reference. A spin-coating speed or time almost always does. These failures are shared, not model-specific. The ranking of fields by difficulty is nearly the same for every generator, and the same small set of fields sits at the bottom for strong and weak models alike: additive and precursor concentrations, the atmosphere of each processing stage and the stability protocol. A stronger model lifts every category but does not remove these modes. Every frontier model knows perovskite chemistry. None of them reliably attaches a value to the device, the step and the condition that produced it, and a stronger reader narrows this gap without closing it. Procedural information, not chemical knowledge, is the part of the record that hampers recovery.

Prompting does not close it either. Switching the annotation Skill on, with everything else fixed, helps every generator (Fig.~\ref{fig:generation}a), but the help is uneven. Fields defined by an encoding convention, such as a flag for sequential deposition\cite{Burschka2013Sequential}, improve sharply. Fields that require the model to align evidence across the paper, such as the list of precursor compounds or the atmosphere of a particular step, can get worse under the same instructions. Policy text tells an agent how to write a value down, but not which device or step the value belongs to. This is why the harness makes verification a separate role rather than trying to prompt its way to reliability. For deployment we chose the generator on the accuracy--cost frontier, a small proprietary model that gives up little accuracy against the best reader for a several-fold saving in cost per paper (Methods).

\subsection*{Verified Harness Brings AI Scientists to Expert Agreement}

\begin{figure*}[!t]
    \centering
    \includegraphics[width=\linewidth]{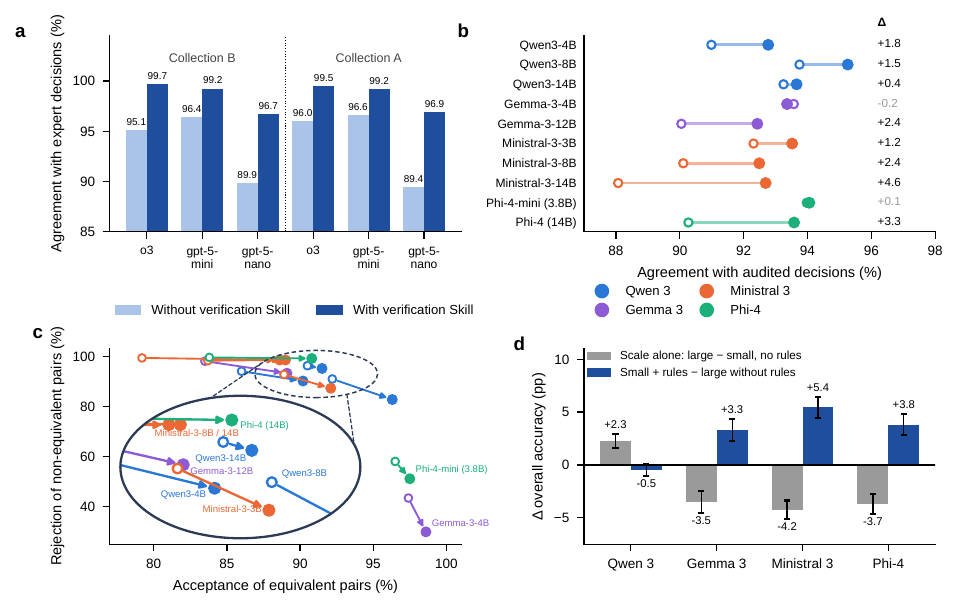}
    \caption{\textbf{Written verification rules bring verifiers to expert agreement.}
    \textbf{a}, Agreement with expert decisions for three proprietary verifiers, with and without the verification Skill, on Collection B ($n$ = 1,460 decisions) and Collection A ($n$ = 1,940).
    \textbf{b}, Agreement of ten open-weight verifiers with the audited o3 decisions (9,700 decisions) without (open) and with (filled) the Skill; $\Delta$, paired difference in percentage points.
    \textbf{c}, Acceptance of equivalent pairs ($n$ = 4,620) against rejection of non-equivalent pairs ($n$ = 548), without (open) and with (filled) the Skill; inset, dashed region enlarged.
    \textbf{d}, Within each family, the largest minus the smallest checkpoint without rules (grey) and the smallest with rules minus the largest without (blue). Error bars, 95\% record-clustered bootstrap intervals (2,000 resamples).}
    \label{fig:verification}
\end{figure*}

The second test moves from the agent that writes the record to the agent that checks it. A verifier is not ground truth, and its authority has to come from somewhere. In the verified harness that source is the written rule, and we tested whether rules, rather than the size of the model behind them, can bring a verifier to agreement with experts. We tested this in three steps: we calibrated proprietary verifiers against expert decisions, ran ten open-weight verifiers with and without the rules, and, within each model family, set the smallest checkpoint with rules against the largest without. Compared with the expert decisions on both collections (Fig.~\ref{fig:verification}a), three proprietary verifiers already agree with the experts most of the time without rules. With the verification Skill, their agreement rises to within a few decisions per thousand, and the gain is largest for the weakest model. The strongest configuration, audited field by field, is the reference verifier used for the generator benchmark above and for the open-weight study that follows.

To see the mechanism, we ran ten open-weight verifiers from four model families on the same field decisions, with and without the Skill, and scored their agreement with the audited reference (Fig.~\ref{fig:verification}b and Supplementary Fig.~\sref{fig:ed-verifiers}). The Skill improves almost every verifier. It pulls the checkpoints closer together and helps most where the starting point is weakest. Most non-empty pairs are identical once trivially normalized and are decided correctly either way. The Skill acts on the remainder, where the candidate and the reference say the same thing in different words, and nearly every decision it changes turns a rejection into an acceptance (Fig.~\ref{fig:verification}c and Supplementary Table~\sref{tab:verifier-acceptance}). The gains land where surface form misleads: journal aliases, synonyms for a deposition method, notation for humidity, units for a scan rate. Two fields regress because of defects in the policy, which the next version of the Skill corrects. Ordered layer stacks improve but remain the hardest decision. The models do not lack chemistry; without rules, a verifier rejects an equivalent value because it is written differently, and the Skill tells it that the two are the same.

Scale does not order verifier reliability (Fig.~\ref{fig:verification}d and Supplementary Table~\sref{tab:scale-vs-skill}). In three of the four families, the larger checkpoint is less accurate without rules than the smaller one, and the smaller checkpoint with the Skill beats the larger one without it, by margins whose bootstrap intervals exclude zero. In the fourth family the largest model without rules still edges the smallest with rules, by a margin that cannot be told from zero. Two checkpoints that the Skill does not help show the limit of what rules can do. They already accept almost every equivalent pair and fail instead by accepting values that are wrong, and rules that license equivalences cannot make a verifier stricter (Fig.~\ref{fig:verification}c). Rules replace scale for verifiers whose errors are lexical rejections, which is the common case, but not for verifiers whose errors are permissive acceptances. The reference labels are the decisions of one proprietary model, so closer agreement may in part mean adopting that model's conventions; the expert audit bounds this risk for the reference verifier itself. In practice, a laboratory can run an expert-grade verifier on a small open-weight model if the rules exist, and it can fix a wrong rule in the next version of the Skill, neither of which a larger model offers. The harness, not the model, is what brings these AI scientists to expert agreement.

\subsection*{Scaling Inference-time Compute Guarantees Reliability}

Reliability at the scale of a field is bought with inference-time compute. After the generator has written a record, every field of it is checked again, by two independent judges of different capability, and the cost of these checks is the price of a measured reliability. A reliability measured on a hundred and fifty papers is useful only if it survives this deployment and is delivered together with the data. We ran the harness over the peer-reviewed experimental PSC literature of the past five years, with the frontier-cost generator chosen above. The judges verified every field of every record under the verification Skill (Methods). Their disagreement is a cheap check on the evaluation itself. The harness returned one record per paper and a judged accuracy for every category of field (Supplementary Fig.~\sref{fig:database}a). The two judges differ by less than one accuracy point in any category.

The pattern across categories matches where the information sits in a paper, and it is the pattern the benchmark predicted. Fields that are local and lexically anchored are near ceiling: the reference data, the cell definition, the fabrication steps and the J--V metrics. Fields that require following a procedure across sections are harder: the device stack and the material synthesis. Stability fields are hardest, because a stability protocol mixes conditions, durations and relative values that are often reported only in a figure or in the supporting information. Recall exceeds precision overall and in every category but one (Supplementary Fig.~\sref{fig:database}b). The generator rarely misses information that is present; its usual failure is an incorrect value, the same profile the benchmark found on the expert-curated papers. Low-precision categories can propagate incorrect values into later analysis, an instance of the downstream effects of data-quality problems documented in other AI pipelines\cite{Sambasivan2021DataCascades}.

The database covers literature through April 2025 and complements the manually curated resource\cite{jacobsson2022perovskitedatabase} and earlier automatically extracted datasets\cite{beard2022perovskite,valencia2025fabricationdb}. Annotating the whole corpus cost a small fraction of the expert time behind the manually curated database. Category reliability is released alongside the records, extending the documentation of intended uses and evaluation conditions advocated by datasheets and model cards\cite{Gebru2021Datasheets,Mitchell2019ModelCards}. These category scores do not provide confidence estimates for individual values. From it follows a rule for anyone who uses the data, person or agent: a field may enter automated analysis only if its category reliability is high enough. Our discovery experiments obey this rule, and neither uses stability, the least reliable category.

\subsection*{From Knowns to Unknowns}

The last test is the use of the records in scientific reasoning, from the knowns of the field to its unknowns. We ran it at three levels: the recovery of trends that the field has established by other means, the prediction of devices that no model has seen, and the use of the records by a team of AI scientists under the reliability rule.

\begin{figure*}[!t]
    \centering
    \includegraphics[width=\linewidth]{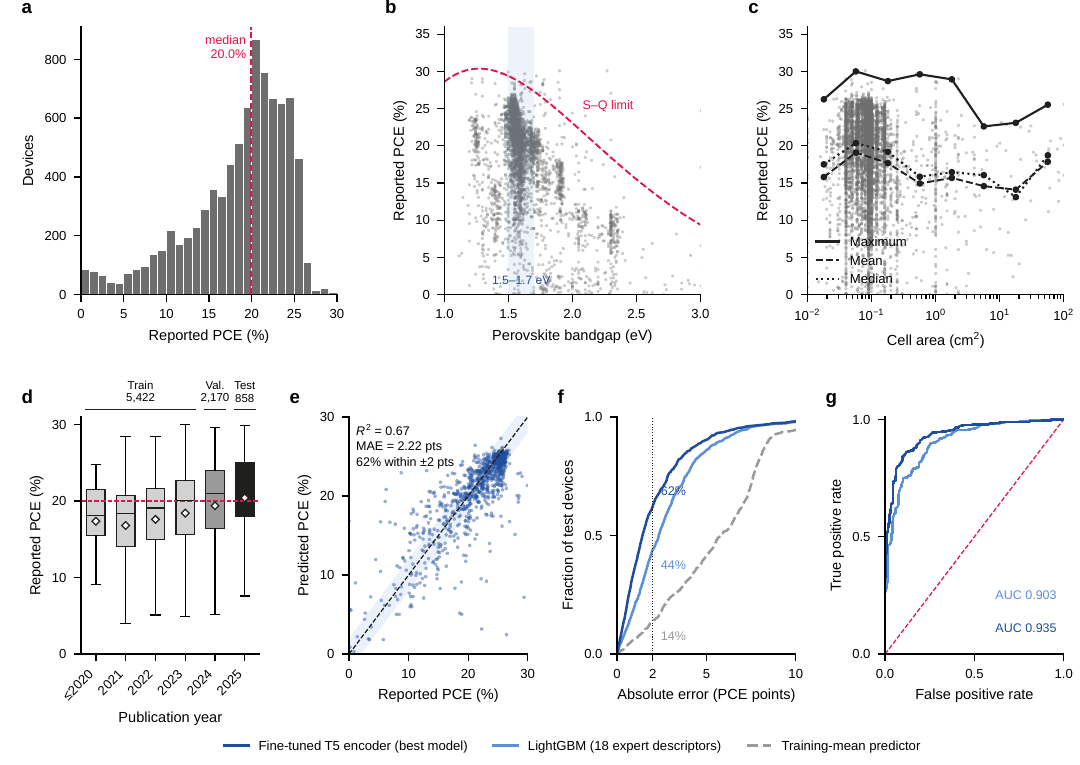}\\[4pt]
    \begin{minipage}{\textwidth}\sffamily\fontsize{8}{8}\selectfont
    \noindent{\bfseries\fontsize{9}{10}\selectfont h}\par\vspace{-1pt}
    \centering
    \begin{tabular}{@{}llcccc@{}}
    \toprule[1.5pt]
    Model & Input & MAE (PCE points) $\downarrow$ & MAD:MAE $\uparrow$ & $R^2$ $\uparrow$ & AUC, PCE $>$ 20\% $\uparrow$ \\
    \midrule
    Training-set Readout & -- & 5.69 & 0.80 & $-$0.22 & 0.500 \\
    \addlinespace[1.5pt]
    LightGBM (18 expert descriptors) & tabular & 2.87 $\pm$ 0.01 & 1.59 $\pm$ 0.01 & 0.60 $\pm$ 0.00 & 0.904 $\pm$ 0.001 \\
    LightGBM (all recipe fields) & tabular & 2.48 $\pm$ 0.01 & 1.84 $\pm$ 0.00 & 0.69 $\pm$ 0.00 & 0.930 $\pm$ 0.001 \\
    TabPFN v2 (all recipe fields) & tabular & 2.49 $\pm$ 0.01 & 1.84 $\pm$ 0.00 & 0.70 $\pm$ 0.00 & 0.932 $\pm$ 0.001 \\
    \addlinespace[1.5pt]
    MatSciBERT (fine-tuned) & text (512 tokens) & 2.24 $\pm$ 0.09 & 2.04 $\pm$ 0.08 & \textbf{0.71 $\pm$ 0.03} & \textbf{0.935 $\pm$ 0.002} \\
    T5-base encoder (fine-tuned) & text (1,024 tokens) & \textbf{2.20 $\pm$ 0.06} & \textbf{2.08 $\pm$ 0.05} & 0.68 $\pm$ 0.02 & 0.934 $\pm$ 0.002 \\
    \addlinespace[1.5pt]
    Llama-3.1-8B embeddings + ridge & text & 2.70 & 1.69 & 0.63 & 0.912 \\
    Llama-3.1-8B-Instruct (zero-shot) & text & 4.10 & 1.12 & $-$0.02 & 0.642 \\
    Llama-3.1-8B-Instruct (five-shot) & text & 2.82 & 1.62 & 0.50 & 0.867 \\
    \bottomrule[1.5pt]
    \end{tabular}
    \end{minipage}
    \captionsetup{font=footnotesize}
    \caption{\textbf{Reliable fields rediscover known trends and predict devices reported a year later.}
    \textbf{a}--\textbf{c}, The 10,258 champion devices read from the high-reliability categories: \textbf{a}, reported PCE (dashed line, median); \textbf{b}, PCE against bandgap with the Shockley--Queisser limit (dashed) and the 1.5--1.7~eV window (shaded); \textbf{c}, PCE against cell area with the maximum, mean and median per logarithmic bin.
    \textbf{d}--\textbf{g}, Forecasting 2025 devices from recipe fields alone: \textbf{d}, reported PCE by publication year and the temporal split (box, interquartile range; whiskers, 1.5$\times$); \textbf{e}--\textbf{g}, the fine-tuned T5 encoder on the 858 test devices, with predicted against reported PCE (band, $\pm$2 points), the cumulative absolute error and the ROC curve for PCE $>$ 20\%; the expert-descriptor LightGBM and the training-mean predictor are shown for reference.
    \textbf{h}, Benchmark on the same split: MAE, the ratio of the mean absolute deviation of the targets to the MAE (MAD:MAE), $R^2$ and the AUC for PCE $>$ 20\%; mean $\pm$ s.d. over three seeds; bold, best per metric.}
    \label{fig:landscape}
\end{figure*}

Figure~\ref{fig:landscape}a--c shows the J--V landscape read from the high-reliability categories alone. The distribution of reported efficiency has the shape the field would expect, and so do the distributions of open-circuit voltage, short-circuit current density and fill factor. High-efficiency devices cluster in a narrow band of bandgaps below the Shockley--Queisser limit\cite{Miah2024BandgapRange,ShockleyQueisser1961} (Fig.~\ref{fig:landscape}b). Efficiency falls and spreads as the device area grows\cite{Yang2024IndustrializationPSC} (Fig.~\ref{fig:landscape}c). Within this corpus, regular (\emph{n--i--p}) devices have higher peak efficiencies than inverted ones; this is a property of the sampled years, not a universal ordering of architectures\cite{Zhang2024PINPerovskiteReview}. Composition and surface-treatment associations are consistent with the broader literature on mixed-cation absorbers and interfacial modification\cite{Byranvand2022MixedASiteReview,Li2023Mixed2D3DReview,Jiang2019Passivation}. Recovering these relationships supports aggregate use of the records, without establishing the accuracy of every value or a causal effect of treatment. Each record is a paper's champion device, so these distributions describe what was reported, not every device that was made.

For an AI scientist, a literature-derived dataset matters only if the knowledge it encodes transfers to devices it has not seen. Following LLM4Mat-Bench's comparison of task-specific and general-purpose predictors\cite{Rubungo2025LLM4MatBench}, we posed a temporally held-out task: predict a device's efficiency from its recipe alone, the fields describing the cell, the stack, the synthesis and the fabrication. Models learn from papers published up to one year, are tuned on the next and are scored once on the year after that. Every performance, stability and bibliographic field is withheld to reduce target leakage (Methods)\cite{Kapoor2023Leakage}. We compared gradient-boosted trees\cite{Ke2017LightGBM} and a tabular foundation model\cite{Hollmann2025TabPFN}, text encoders fine-tuned on a plain-text serialization of the recipe\cite{Gupta2022MatSciBERT,Raffel2020T5,Rubungo2023LLMProp}, and a general-purpose LLM\cite{Grattafiori2024Llama3} whose training data end before any test paper appeared, used either through frozen embeddings or prompted with retrieved examples (Fig.~\ref{fig:landscape}h). Every trained model clearly beats the predictor that always guesses the training mean, even though reported efficiencies drift upward from the training years to the test year. A tree model on a short list of expert-chosen descriptors already separates good devices from poor ones. Giving it every recipe field improves it further, so the full schema that the harness reconstructs carries information that expert descriptors leave out. The fine-tuned text encoders, which read the record as a few lines of text and need no feature engineering, do best of all, and most of their remaining error sits on low-efficiency devices (Fig.~\ref{fig:landscape}d--g; every model is shown in Supplementary Figs.~\sref{fig:pred-tabular} and \sref{fig:pred-text}). With the architecture held fixed, formamidinium content and bandgap rank highly among the descriptors. Studies of composition and formamidinium absorbers make these associations plausible\cite{Jeon2015NatureCompositional,Min2019ScienceFAPbI3}, but feature importance alone does not establish a mechanism. A general-purpose LLM is no substitute for training on the records: prompted without examples it predicts nearly the same efficiency for every device, and the examples it is handed by retrieval account for most of what it gains. Residual error may reflect omitted processing choices, laboratory differences, extraction errors and model limitations; measurement conventions are another documented source of variability\cite{Christians2015Measurements}.

\begin{figure*}[!t]
    \centering
    \includegraphics[width=\linewidth]{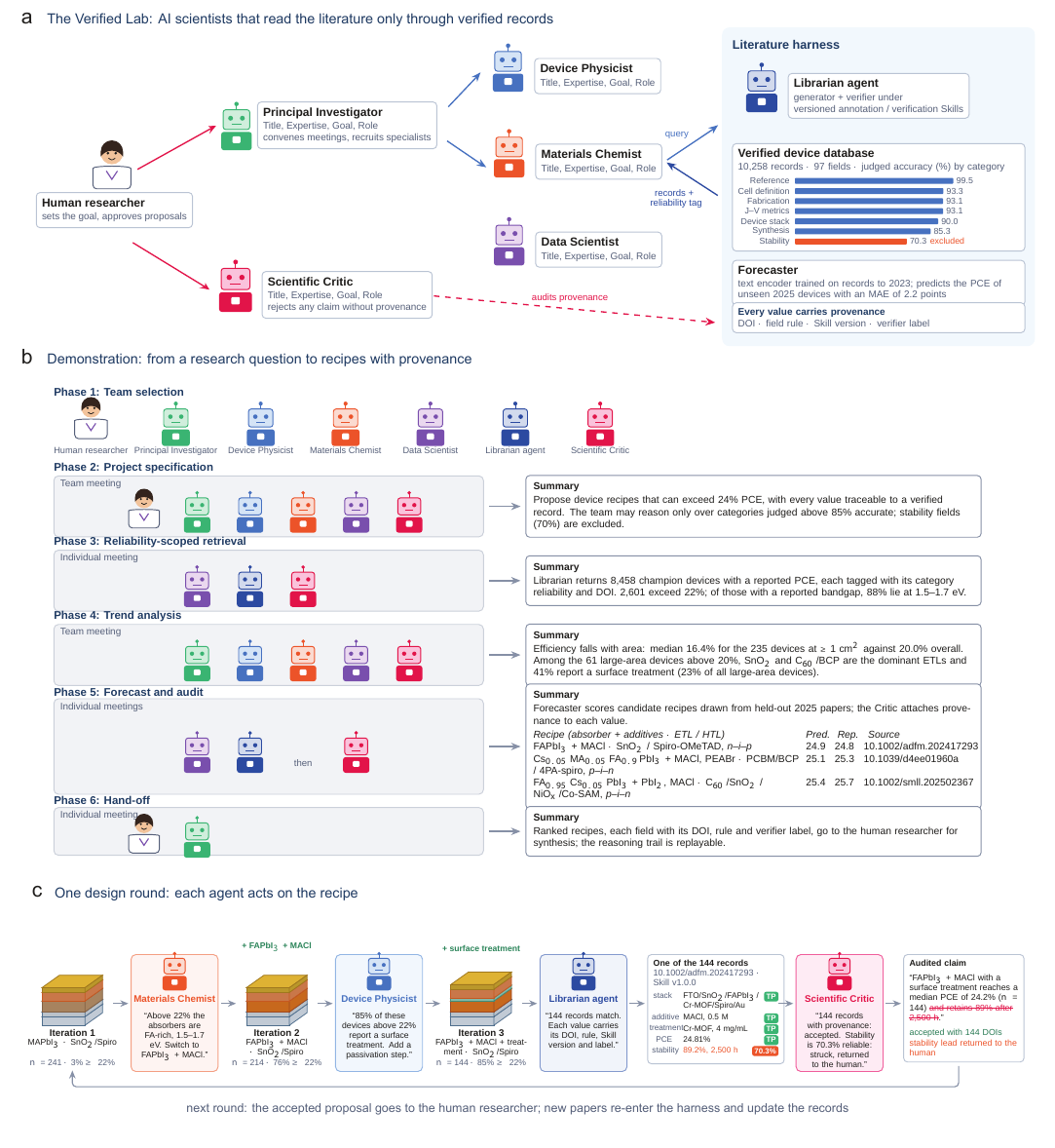}
    \captionsetup{font=footnotesize}
    \caption{\textbf{A team of AI scientists reasoning over verified records.}
    \textbf{a}, Architecture, after the Virtual Lab\cite{Swanson2025VirtualLab}. A human sets the goal; a PI agent recruits specialists and a critic; every agent reaches the literature through the librarian, which returns records tagged by category reliability and provenance, and a forecaster trained on records to 2023 scores candidate devices.
    \textbf{b}, Six phases of a design question. Dialogue is illustrative and abridged; all numbers are computed from the released records and the forecaster. The recipes in Phase 5 are held-out 2025 devices chosen for illustration.
    \textbf{c}, One design round: the device at each iteration, the agent acting on it, the record the librarian returns with its verifier labels (TP, true positive), and the claim after the critic strikes the clause that rests on an unreliable category. $n$ and the share above 22\% are computed over the matching verified records (Methods).}
    \label{fig:verifiedlab}
\end{figure*}

At the third level we sketched the Verified Lab, a team of LLM agents modelled on the Virtual Lab\cite{Swanson2025VirtualLab} in which every agent reaches the literature only through the harness (Fig.~\ref{fig:verifiedlab}a). A principal-investigator agent plans the study and convenes three specialists (a device physicist, a materials chemist and a data scientist) and a scientific critic. A librarian agent, which is the harness itself, answers their queries with records that carry their category reliability and their provenance, and computes the evidence directly from those records. A reliability gate admits only the categories above the threshold, and the critic checks each drafted claim against the evidence before the harness releases or withholds it. A forecaster, the text encoder trained above, scores candidate recipes. In a six-phase walk-through of a design question (Fig.~\ref{fig:verifiedlab}b), the team confines its reasoning to the categories the gate admits. It retrieves the champion devices with a reported efficiency, recovers that the best of them sit in a narrow bandgap window and that efficiency falls with area, and returns recipes from held-out papers whose forecast efficiencies lie close to the reported values. Each field is linked to its DOI, its rule and its verifier label. One design round is laid out as a pipeline (Fig.~\ref{fig:verifiedlab}c). Two specialists move a recipe from a single-cation absorber to a formamidinium-based one with an additive and a surface treatment. The librarian returns the matching records with their labels. The critic strikes the one clause that rests on an unreliable category. The dialogue is illustrative, but every number in it is computed from the released records. The demonstration shows the contract we propose for AI scientists, not a validated discovery; its running implementation, a research workspace released with the code, is shown in Supplementary Figs.~\sref{fig:ws-overview}--\sref{fig:ws-skills}. The same abstractions serve a human laboratory and an AI scientist (Fig.~\ref{fig:overview}b): a typed record in place of the PDF, a versioned Skill in place of a model-specific prompt, a reliability-scoped query in place of recall, and a verifier call and a provenance call in place of the colleague who re-derives a number by hand. Together, these results indicate that the verified records carry the knowns of the field in a form from which its unknowns can be approached, at all three levels, and that the reliability rule is what makes each level credible. The trends are read from categories whose accuracy is known, the forecaster is trained on them, and the critic enforces them.

\section*{Discussion}

We set out to carry the experimental literature across its newest boundary with its meaning intact, and to measure the agents that do the carrying. Separating the writing of a record from its checking, governing both with written scientific policy and holding everything else fixed turned the output of a language model into a measured artefact. The result is a benchmark that shows where research agents fail, a verifier whose agreement with experts is known, and a field-wide database in which every category of field carries a judged accuracy. The same error profile appears on the benchmark and in the deployment. Agents recover isolated facts, fail when a value has to be tied to a device, a step or a condition, and fail by writing a plausible wrong value rather than by leaving a blank. These are the errors that an aggregate accuracy hides and that pass silently into an AI scientist's plans.

Three principles follow, and each runs against a common assumption about LLM-driven science. First, the bottleneck is binding, not knowledge. Existing systems demonstrate scientific text extraction and tool-augmented reasoning\cite{Dagdelen2024StructuredExtraction,Polak2024ChatExtract,Bran2024ChemCrow}. These successes leave a separate question: whether each extracted value remains attached to the correct experiment. Every frontier model fails in the same procedural places. A stronger model narrows the gap without closing it, and prompting fixes conventions but not alignment. Second, rules are worth more than scale, and rigour should be written down rather than hoped for. A verifier's authority can come from calibration against experts under explicit, versioned rules rather than from the size of its backbone. The gain falls on exactly the decisions that string matching cannot settle, and the defects the audit exposed were fixed in the next version of the Skill. Writing the reviewing consensus of experts, the equivalence rules of chemistry and the boundary conditions of physics into versioned Skills makes verification inspectable, correctable and portable across models, none of which a larger model provides. The limit of rules is equally clear: they license equivalences, so they repair verifiers that are too strict and not verifiers that are too lenient. Third, reliability travels with the data. A reliability measured once on a benchmark and then set aside protects no one. Measured per category and attached to every record, it becomes the rule by which any later analysis, human or automated, decides what to trust. Seen against the history we began with, these principles are the rules for the newest boundary. They do for the machine-read record what review and citation did for print.

The harness has limitations that we consider inherent in current LLM-based pipelines, and each points to a remedy. The reliability of the full database is judged by language models rather than by complete expert annotation, and one judge shares the generator's backbone and may favour its output\cite{Panickssery2024SelfPreference}. The expert audit bounds this risk for the calibrated verifier and the disagreement between the two judges keeps it visible, but a sample of deployed records should be re-audited as the database is reused. The verification Skill describes equivalences in more detail than non-equivalences, which is why rules help over-strict verifiers and not lenient ones; explicit guardrails against false acceptance are the next addition. We studied one materials domain, so the schema and category scores need not transfer. Documenting the corpus and its intended use is necessary before generalizing to another setting\cite{Bender2018DataStatements}. Each record is a paper's champion device and positive results are more likely to be published, so the database describes reported best devices. The trends and forecasts are associations tested on one held-out year; they suggest hypotheses and do not establish causes, consistent with cautions about interpreting data-driven materials models\cite{Butler2018MLMaterials,Schmidt2019MLSolidState}. The weakest categories, synthesis and stability, require following a procedure across a long document and reading figures and supplements. Multimodal inputs and targeted human review of these fields are the most direct ways to close the gap. Finally, the harness relies on commercial models whose versions change; we pin every version and release the records as the reference artefact.

Human expertise remains part of the design. Experts write the schema, author and version the Skills, adjudicate the reference records and audit the verifier. The harness scales their judgement to a whole field and reports, category by category, how far that scaling can be trusted. The error profile suggests a division of labour that we have motivated but not yet tested end to end. Localized scalars and exact matches can be accepted automatically, conditions and relational fields need verification under calibrated rules, and ordered sequences, multi-part concentrations and stability protocols stay with the experts.

Looking ahead, just as journals now expect data and code to be available, we anticipate that research agents will expect the literature to arrive as verified records with an explicit reliability. The ease with which a paper can be turned into such a record will itself become a measure of how reproducibly it was written. Every earlier medium of knowledge acquired its conventions after the fact, once the cost of unprotected transmission had become visible. A verified-literature harness offers AI scientists in the experimental sciences those conventions in advance: the data they reason over, and a measure of how far that data can be trusted. As agents move from answering questions to choosing what to synthesize and measure, that measure becomes a prerequisite.

\section*{Methods}

\subsection*{Harness design}

\paragraph{Task.} Given a paper package $d$ (main text plus supporting information) and a schema $\mathcal{S}=\{s_j\}_{j=1}^{m}$ of $m=97$ fields, the harness outputs one record $r(d)$ describing the paper's best-performing device, with every field filled by a string value or an explicit \texttt{Null}. Selecting one device per paper gives a comparable unit across heterogeneous reports; it also introduces a selection effect that we account for in the discovery analyses. The 97 fields fall into seven categories (reference data, cell definition, device stack, material synthesis, device fabrication, J--V metrics and stability metrics).

\paragraph{Versioned Skills.} Scientific policy is separated from the code that executes it. The annotation Skill $A^{\alpha}=(\mathcal{S},\mathcal{C})$ (released as \texttt{psc-annotate-v1@1.0.0}) couples the schema with a policy $\mathcal{C}$ for champion-device selection (rank by explicitly reported scan PCE, not stabilized output, certification or module efficiency), evidence priority (prefer explicit device-specific statements; map sample codes across article and supplement before combining fabrication and performance values), layer alignment (report the complete electrically active stack; align slash-delimited properties with layer order), measurement identity and the use of \texttt{Null} (when a value is absent, ambiguous between devices or only inferable). The verification Skill $V^{\beta}=(\mathcal{R},\mathcal{L})$ (\texttt{psc-verify-v1@1.1.0}) supplies field rules $\mathcal{R}=\{r_j\}$ defining chemical aliases, compatible units, numerical tolerances, ordered-sequence semantics and device-specific comparisons, together with the allowed label set $\mathcal{L}$. A machine-readable registry pins the full semantic version used in every run; model changes preserve the evaluation contract, whereas changes to a schema, tolerance or decision rule require a new Skill version. The model-facing views materialized from both Skills are released with the code.

\paragraph{Execution boundary.} The harness $H$ parses the paper package, materializes the annotation Skill for a model adapter, validates candidate keys and types (rejecting malformed records), pairs each candidate value with the expert reference and field rule into $q_j=(s_j,\hat{y}_j,y^*_j,r_j)$, invokes the verifier on each unit, validates the returned label against $\mathcal{L}$, aggregates metrics and writes an audit bundle containing document, candidate, reference, policy versions, model versions and decision provenance. Every generator and verifier traverses this same boundary.

\paragraph{Generator and verifier agents.} A generator agent reads the full paper package, identifies the best-performing device and emits a JSON record conforming to $\mathcal{S}$. A verifier agent receives, for each field, the candidate value, the reference value and the field rule and returns one of TN (both empty), FP (reference empty, candidate filled), FN (reference present, candidate empty), TP (both present and scientifically equivalent) or FP$_{\mathrm{mismatch}}$ (both present and non-equivalent). Accuracy, precision, recall and $F_1$ follow the standard definitions with the false-positive count equal to $\mathrm{FP}+\mathrm{FP}_{\mathrm{mismatch}}$. Under this metric an entirely empty record scores 36.4\% accuracy on Collection A, which we report as the empty-record baseline.

\subsection*{Expert-curated testbed}

The reference comprises two collections. Collection A contributes 100 papers annotated by us against all 97 fields; Collection B contributes 50 papers with 73 overlapping fields drawn from the Perovskite Database\cite{jacobsson2022perovskitedatabase}. For each paper, domain experts inspected the article and supplement, annotated independently against the pinned annotation Skill, identified field-wise disagreements and resolved them by discussion and re-inspection before any automated verifier was evaluated. The collections total 13,350 field decisions; an average paper package is 23,231 input tokens and an average record 1,586 output tokens.

\subsection*{Generator benchmark}

We evaluated 16 proprietary and 2 open-weight generators (release date, pinned version and context length). Each generator produced one record per Collection A paper with and without the annotation Skill, with the harness, schema, source text, parser and output validator held fixed; the o3 verifier with the verification Skill, calibrated as described below, provided the field decisions. We report accuracy, precision, recall and $F_1$ over 9,700 decisions per generator (Supplementary Fig.~\sref{fig:ed-generators}) and accuracy by subfield (Supplementary Fig.~\sref{fig:ed-subfields}). Error composition decomposes decisions into omissions (reference present, candidate \texttt{Null}), wrong values (both present, non-equivalent) and over-prediction (reference \texttt{Null}, candidate filled), each conditioned on the relevant reference state so that subfields with different null shares remain comparable. The three field classes (localized scalar, 24 fields; relational, 30; condition, 43) are an analysis grouping of the schema by the competence a field exposes, assigned from the schema definitions, not labels in the dataset. Per-field accuracy ranks were compared across generators with pairwise Spearman correlation. Request cost per paper for each proprietary generator was taken from the provider's list price at the pinned version; the two open-weight generators have no API price and are shown without cost. Annotation-Skill variants with increasing field-specific detail were materialized for gpt-5-mini and gpt-4o-mini to measure how much domain policy contributes beyond the base model; these are experimental variants, not interchangeable versions of the released Skill.

\subsection*{Verifier calibration against experts}

Three proprietary backbones (o3, gpt-5-mini and gpt-5-nano) verified generator records against the expert references on both collections, with and without the verification Skill, and each automated label was compared with the expert decision for the same field (Fig.~\ref{fig:verification}a). The o3 configuration with the Skill was then audited field by field and used as the reference verifier for the generator benchmark and the open-weight study.

\subsection*{Open-weight verifier study}

Ten open-weight checkpoints (Qwen3-4B, -8B, -14B; Gemma-3-4B, -12B; Ministral-3-3B, -8B, -14B; Phi-4-mini (3.8B), Phi-4 (14B)) verified the same deterministic sample: a paper-round-robin pass selecting one evaluated generator record from each of the 100 Collection A papers, giving 9,700 field decisions per condition (Supplementary Fig.~\sref{fig:ed-verifiers}). Audited o3 decisions with the Skill provided the comparison labels (4,622 TP, 3,107 TN, 548 FP$_{\mathrm{mismatch}}$, 323 FP and 1,100 FN). Candidate--reference missingness assigns TN, FP or FN before inference, so the verifier decides only the 5,198 non-empty comparisons (TP versus FP$_{\mathrm{mismatch}}$); the other 4,502 fields form the missingness stratum, whose accuracy (0.9636) is identical in both conditions. We report overall accuracy over all 9,700 decisions and semantic accuracy over the 5,198 non-empty decisions. All runs used temperature 0, seed 42, bfloat16 inference, a 16,384-token context and JSON-schema-constrained decoding through xgrammar in vLLM 0.12.0\cite{Kwon2023PagedAttention} on NVIDIA A100 40GB GPUs, with Qwen3 thinking disabled; each run configuration pins model and tokenizer revisions, prompt hashes, source digests, package versions and GPU metadata, and all 200 outputs per backbone were recoverable with no invalid record or field. The Skill effect is the paired difference $\Delta=\mathrm{Acc}_{\mathrm{with}}-\mathrm{Acc}_{\mathrm{without}}$; we resampled the 100 records with replacement, retaining all 97 fields within each sampled record, and report 2.5th and 97.5th percentiles over 200 resamples for the per-model deltas and over 2,000 resamples for the within-family scale-versus-rules comparisons (Supplementary Table~\sref{tab:scale-vs-skill}). Acceptance and rejection rates (Supplementary Table~\sref{tab:verifier-acceptance}) are accuracies on the non-empty pairs the reference labels TP and FP$_{\mathrm{mismatch}}$, respectively. The field-wise audit pools each field across the ten checkpoints.

\subsection*{Corpus and deployment}

We collected 10,258 peer-reviewed experimental PSC papers published between 2021 and April 2025 and indexed in Web of Science, each as main text plus supporting information. gpt-5-mini (\texttt{gpt-5-mini-2025-08-07}) served as the generator under the annotation Skill; it was chosen because it lies on the accuracy--cost frontier of the generator benchmark. Every record was then judged field by field by two independent verifier agents, gpt-5-mini and o3 (\texttt{o3-2025-04-16}), under the verification Skill. We report accuracy, precision, recall and $F_1$ per field category as the mean over the two judges, with their standard deviation as a check on the evaluation itself. The released records carry per-record provenance linking each record to its source digital object identifier (DOI), and category-level reliability scores. The benchmark builds on MavaLM, published as a NeurIPS 2026 AI4Science workshop poster\cite{Dong2026MavaLM}.

\subsection*{Reliability-scoped discovery}

Following the reliability rule, neither discovery analysis uses stability fields. The rediscovery analysis reads the high-reliability categories (cell definition, device stack, device fabrication and J--V metrics); the predictive models additionally read material-synthesis fields as inputs. Distributions in Fig.~\ref{fig:landscape} use all records with a parsable value for the plotted field; the Shockley--Queisser curve in Fig.~\ref{fig:landscape}b is the detailed-balance limit for a 5,778~K blackbody Sun normalized to 1,000~W\,m$^{-2}$ and a 300~K cell.

\subsection*{Qualitative demonstration}
The Verified Lab (Fig.~\ref{fig:verifiedlab}) is a design sketch of how AI-scientist agents should consume the records, not an experiment; the agents, meetings and summaries are illustrative. Its statistics use the same parsing as Fig.~\ref{fig:landscape}: records with a parsable reverse-scan PCE in (0, 30], bandgap from the perovskite bandgap field and cell area from the measured-area field (mm$^2$ converted to cm$^2$). Large-area devices are those with an area of at least 1~cm$^2$; dominant electron-transport layers are the most frequent stack sequences among large-area devices above 20\% PCE; a surface treatment is counted when the surface-treatment material field is not \texttt{Null}. Forecasts are those of the fine-tuned T5 encoder (seed with the median test MAE). The three recipes shown were drawn from test-set (2025) devices with non-empty composition, electron- and hole-transport fields and an absolute forecast error below 0.7 PCE points, so they illustrate the output format rather than typical accuracy, which Fig.~\ref{fig:landscape}d--h report. In the design round of Fig.~\ref{fig:verifiedlab}c, MA-only absorbers are records whose composition names MAPbI$_3$ (or CH$_3$NH$_3$PbI$_3$) without FA or Cs and whose electron- and hole-transport stacks contain SnO$_2$ and Spiro-OMeTAD; FAPbI$_3$ recipes are compositions beginning with FAPbI$_3$ without Cs or MA, with MACl named in the additive field; a surface treatment is any non-\texttt{Null} surface-treatment material. The record shown there is reproduced from the database.

\subsection*{Temporal prediction benchmark}

\paragraph{Data and split.} The target is the reverse-scan PCE, parsed as the first number in the field. We kept the 8,450 records with a publication year and $0<\mathrm{PCE}\le30\%$, which removes tandem devices and garbled values. Records from papers published up to 2023 form the training set (5,422), 2024 the validation set (2,170) and 2025 the test set (858). The validation set was used only for early stopping and for choosing the embedding layer and ridge penalty; the test set was scored once. The share of devices above 20\% PCE rises from 43.5\% (train) to 58.2\% (validation) and 65.5\% (test).

\paragraph{Inputs.} Models see the 71 recipe fields of the schema (cell definition, device stack, synthesis and fabrication). Bibliographic and provenance fields, all J--V fields and all stability fields were removed, because PCE equals $V_\mathrm{OC}\,J_\mathrm{SC}\,\mathrm{FF}/P_\mathrm{in}$, forward-scan, stabilized and certified PCE nearly copy the target, and 610 titles contain a percentage, usually the efficiency.

\paragraph{Tabular features.} The expert descriptors are 18 variables: presence of formamidinium (FA), methylammonium (MA) and Cs on the A site, Pb and Sn on the B site, and I, Br and Cl on the X site, parsed from the composition string; perovskite bandgap; absorber thickness; measured cell area; whether a perovskite additive, a surface treatment and an electron-transport-layer (ETL) additive are reported; substrate (ITO, FTO, other); ETL deposition method (spin coating, chemical bath, vapour phase or atomic layer deposition, other); hole-transport-layer (HTL) deposition atmosphere (N$_2$, air, other); and sequential spin coating. For all recipe fields, fields with fewer than 10 training values were dropped; a field was treated as numeric if at least 60\% of its training values contain exactly one number and was mapped to that number; otherwise its training values with at least 10 occurrences became categories and the rest mapped to ``other''. The eight composition-site flags were appended, giving 77 features.

\paragraph{Models.} LightGBM used a learning rate of 0.03, 31 leaves, row and column subsampling of 0.8 and early stopping on validation MAE (200 rounds). TabPFN v2 (v2.1.3) used default settings on the all-field table. Text models read a deterministic serialization of the non-empty recipe fields as ``field name: value'' lines, absorber fields first, built without any LLM. MatSciBERT (512 tokens) and the T5-base encoder (1,024 tokens) were fine-tuned with mean pooling, a linear head and L1 loss on the standardized target for 10 epochs (AdamW\cite{Loshchilov2019AdamW}, batch size 16, learning rates $5\times10^{-5}$ and $10^{-4}$, 10\% warm-up), keeping the epoch with the lowest validation MAE. For the frozen-embedding baseline, mean-pooled hidden states of Llama-3.1-8B-Instruct from layer 16 or 32 fed a ridge regression, with layer and penalty chosen on validation. Prompted Llama-3.1-8B-Instruct decoded greedily for up to 48 tokens; five-shot prompts added the five training records most similar to the query under TF-IDF. An answer was valid only if it was a bare number or ended with a statement of the form ``\ldots to be $X$\%''; invalid answers were scored as the training mean and their rate is reported.

\paragraph{Metrics and statistics.} We report MAE; the ratio of the mean absolute deviation of the test targets to the MAE (MAD:MAE, as in LLM4Mat-Bench\cite{Rubungo2025LLM4MatBench}; values above 1 beat predicting the mean); $R^2$; and the ROC AUC for PCE $>20\%$ computed with the predicted PCE as the score. Stochastically trained models were run with three seeds and are reported as mean $\pm$ s.d. Significance between models uses a one-sided paired bootstrap\cite{Efron1979Bootstrap} on test MAE (10,000 resamples) over seed-averaged predictions.

\paragraph{Controlled-subset analysis.} As a complementary analysis holding the device architecture fixed, we modelled the 2,445 devices with an SnO$_2$ ETL and a Spiro-OMeTAD HTL\cite{Jiang2016SnO2,AvilaLopez2025SpiroOMeTADReview}, described by the 18 expert descriptors, with 5-fold cross-validation and grid-searched hyperparameters within each fold. Regression models were gradient boosting, XGBoost\cite{Chen2016XGBoost}, LightGBM, random forest\cite{Breiman2001RandomForest}, extra trees, decision tree, AdaBoost and $k$-nearest neighbours; classification models (PCE $>$ 20\%) were random forest, LightGBM, XGBoost, gradient boosting, extra trees and decision tree. Feature importance is gain-based for boosted and impurity-based for bagging ensembles; importance rankings from any single model should be read as hypotheses, not mechanisms.

\section*{Data availability}
The verified device database (10,258 records, 97 core fields, per-record provenance to the source DOI and category-level reliability scores) is released as Parquet and gzipped JSON Lines files on the Hugging Face Hub at \datasetURL{} and in the \texttt{dataset/} directory of \codeURL{}, together with the discovery code and per-run predictions. The expert-curated reference records, versioned annotation and verification Skills, model settings and field-level verification decisions are released with the code. Only structured factual values with provenance are released, not article text; full reproduction of the harness also requires model credentials and licensed article files.

\section*{Code availability}
The harness runtime, the agentic research workspace, the prediction benchmark (with per-run predictions and raw LLM outputs) and the figure sources are available at \codeURL{}; the manuscript source is kept separately at \url{https://github.com/jiayijdong/Harnessing-Verified-literature-for-AI-scientist}. A two-minute video of the VerifiedLab research workspace, recorded from the running system on local open-weight models, is available at \href{https://youtu.be/P7We1Q_Xbyo}{\texttt{youtu.be/P7We1Q\_Xbyo}}.

\section*{Disclosure of Generative AI Usage}
We used large language models to assist with writing clarity and grammar, with experimental, figure and analysis code, and with drafting parts of the appendix. All final wording, technical claims, code, results, figures’ scientific content, and conclusions were reviewed, validated, and produced by the authors.

\section*{Ethics statement}
The database is built from peer-reviewed publications and does not involve human subjects or personal data. LLM-extracted scientific data can contain confident but incorrect values, and our results show that incorrect values, not omissions, are the dominant error. Users should consult the per-category reliability scores before relying on any field and should not treat model predictions as substitutes for experimental validation or safety assessment.

\bibliographystyle{naturemag-linked}
{\footnotesize\hypersetup{urlcolor=black}\bibliography{references}}

\end{document}